\documentclass[11pt,letter]{article}
\usepackage[utf8]{inputenc}
\usepackage[all]{xy}
\usepackage{amsmath}
\usepackage{amssymb}
\usepackage{mathrsfs}
\usepackage{euscript}
\usepackage{mathdesign}
\usepackage{amsthm}
\usepackage{bm}
\usepackage{comment}
\usepackage{framed}
\usepackage{geometry}
\usepackage{blindtext}
\usepackage{xcolor}
\usepackage{comment}
\usepackage{float}
\usepackage{graphicx, wrapfig}
\usepackage{subfig}

\begin{document}

\theoremstyle{plain}
\newtheorem{thm}{Theorem}[section] % reset theorem numbering for each chapter
\theoremstyle{definition}
\newtheorem{defn}[thm]{Definition} % definition numbers are dependent on theorem numbers
\newtheorem{proposition}[thm]{Proposition} 
\newtheorem{theorem}[thm]{Theorem} 
\newtheorem{cor}[thm]{Corollary} 
\newtheorem{lemma}[thm]{Lemma}
\newtheorem{example}[thm]{Example} %same for example numbers
\newtheorem{remark}[thm]{Remark}

\renewcommand{\qedsymbol}{$\blacksquare$}
\newcommand{\norm}[1]{\left\lVert#1\right\rVert}
\let\oldle\le
\let\le\leqslant
\let\oldge\ge
\let\ge\geqslant
\let\oldemptyset\emptyset
\let\emptyset\varnothing
\let\subset\propersubset
\let\subset\subseteq
\let\epsilon\stupidepsilon
\let\epsilon\varepsilon
\newcommand{\onto}{\twoheadrightarrow}
\newcommand{\hto}{\stackrel{\mu^*}{\longrightarrow}}

\title{Mathematical Perspective of Machine Learning}
\author{Yarema Boryshchak, PhD}
\maketitle

\begin{center}
	Abstract
\end{center}

We take a closer look at some theoretical challenges of  Machine Learning as a function approximation, gradient descent as the default optimization algorithm, limitations of fixed length and width networks and a different approach to RNNs from a mathematical perspective. 

\vspace{5mm}

\section{Introduction} 

In the nutshell the idea of training a neural network (\textbf{NN}) is equivalent to the problem approximation of a given function, $\bm{\mathfrak{f}}$, with the domain, $\bm{\mathcal{D}}$, and codomain, $\bm{\mathcal{C}}$,
\begin{equation}
\label{eq:modnn1}
\mathfrak{f} : \mathcal{D}\to \mathcal{C}
\end{equation}
which depends on some data  of size $N\in \mathbb{N}$ of $k$-dimensional input vectors,  $\bm{\vec{x}_j}\in \mathcal{D}\subset \mathbb{R}^k$ and $l$-dimensional output (label) vectors,  $\bm{\vec{y}_j}\in \mathcal{C}\subset \mathbb{R}^l$, by a composition of functions of the form
\begin{equation}
\label{eq:modnn2}
\vec{P}_i\left(\vec{z}_{i-1}\cdot \vec{w}_i\right) = \left(P_i\left(\vec{z}_{i-1}\cdot \vec{w}_i\right), ...  \, ,  P_i\left(\vec{z}_{i-1}\cdot \vec{w}_i\right) \right),
\end{equation}
where $\bm{P_i}$ is called an activation function of layer $i$, $\vec{z}_{i-1}$ is an output vector of the layer $i-1$, and $\bm{\vec{w}_i}$ is called the weight vector of layer $i$. Once the size of each layer and the choice of each activation function is made, one usually uses, so called, back propagation algorithm, adjusting the values of each weight vector according to some type of gradient descent rule. In other words, one is trying to solve an optimization problem, minimizing the "difference norm" 
\begin{equation}
\label{eq:modnn3}
\bm{\mathfrak{C}} := \norm{f-\tilde{f}}_{d}, \qquad \text{where } \quad \bm{\tilde{f}} = \vec{P}_r\left(\vec{P}_{r-1}(\dots \vec{P}_1)\right)
\end{equation}
and $r\in \mathbb{N}$ is the number of layers of the neural network.

So apriori, we are making a choice of the function $\tilde{f}$  of weights $\vec{w}_1, \dots,  \vec{w}_r$. Note that the dimension of each vector $\vec{w}_i$ is the size of the layer $i$. To simplify the problem we may always find the maximum, $m$, of the size layers, and assume that each $\vec{w}_i \in \mathbb{R}^m$. We are implicitly assuming that for each $i=1, ... , r$, $\vec{P}_i(\vec{0})=\vec{0}$.
\footnote[1]{We shall return to the importance of this condition later to discuss the design and structure of NN}

\vspace{5mm}

\section{Existence of function $\mathfrak{f}$ and the toll of  cost function $\mathfrak{C}$}

First thing to consider, given a labeled data set $\left\{ (\vec{x}_j, \vec{y}_j) : \, j=1, \dots, N \right\}$, if there is a representation function $\mathfrak{f}$ such that $\mathfrak{f}(\vec{x}_j)=\vec{y}_j$ for all $j=1, \dots , N$. This important step is often overlooked in practice. In theory, if there is
\begin{equation}
\label{eq:modnn4}
\vec{x}_j=\vec{x}_k \qquad \text{such that} \qquad \vec{y}_j\ne \vec{y}_k
\end{equation}
then no such function exist. In other words $\mathfrak{f}$ is a function of more variables than provided in the data set and the idea  approximation by $\tilde{f}$ is meaningless. 

One may always assume some measurement error $\epsilon>0$ (noise) of the data set and consider instead a weaker condition
\begin{equation}
\label{eq:modnn5}
\vec{x}_j=\vec{x}_k \quad \implies \quad \norm{\vec{y}_j - \vec{y}_k}_{l^2}<\epsilon
\end{equation}
necessary for existence of a function $\mathfrak{f}$. In such case one may look at the  average values of duplicate points
\begin{equation}
\label{eq:modnn6}
f_{ave}= \frac{\sum_{\vec{x}_j=\vec{x}_k} \vec{y}_k}{ \sum_{\vec{x}_j=\vec{x}_k} 1}
\end{equation}
and choose to approximate $f_{ave}$ instead of $\mathfrak{f}$. There is no guarantee that a good approximation $\tilde{f}$ of function $f_{ave}$ is a good  approximation of $\mathfrak{f}$ itself.

One should also consider the norm $\norm{\cdot}_d$ of the approximation  function $\mathfrak{f}$. It is well known that various classes of "nice" functions are dense in $L^p$ spaces. In particular, the class of functions $\tilde{f}$ defined in (\ref{eq:modnn3}) are  dense with respect to convergence in measure and $L^p$ norm (see \cite{hornik}). This important fact implies that given any functions $\mathfrak{f}$ and any $\epsilon>0$, there is a function $\tilde{f}$ s.t.
\begin{equation}
\label{eq:modnn7}
\norm{\mathfrak{f}-\tilde{f}}_{L^p} < \epsilon.
\end{equation}
 
The approximation function $\tilde{f}$ is a function of weights vectors $\vec{w}_i$. The pursuit of such function $\tilde{f}$ is a two part problem. The first part, defining the structure of the neural network, is done by a human. The methodology behind the choice of NN structures is  at the stage of experimental science. The second part,  weights optimization, is done by a computer, usually capable of trillions of operations per second. Needless to say that the effectiveness of  latter part depends heavily on the former.

 In practice one usually does not use an approximation with respect to $L^p$ norm. Computationally one may only evaluate the function $\mathfrak{f}$ at finitely many points and approximate it by $\tilde{f}$ at such points, often with respect to the $\norm{\cdot}_{l^p}$ norm. Since for all $0<p<q<\infty$,
\begin{equation}
\label{eq:modnn22}
\norm{ \mathfrak{f}(y_j)- \tilde{f}(y_j)}_{l^q}\le \norm{ \mathfrak{f}(y_j)- \tilde{f}(y_j)}_{l^p},
\end{equation}
one can choose any $p>0$ to obtain the approximation for all $q\ge p$. 
\footnote[1]{Since for any sequence $x=\{x_i\}_{i\in\mathbb{N}}$ of complex numbers $x_i$, \, $\norm{x}_{l^2}\le \norm{x}_{l^1}$, \textit{"l1 weights regularization"} is also an \textit{"l2 regularization"}, so \textit{"(l1 and l2)-regularization"} is redundant.}

Here is an interesting question. Given $f\in L^p(\mathbb{R}^k)$, does it follow that the sequence $\{f(\vec{y}_j)\}_{j=1}^{\infty} \in l^p$ for any choice $\vec{y}_j \in \mathbb{R}^k$? One can easily show  it is not the case. What about a sequence of randomly chosen points $\vec{y}_j\in \mathbb{R}^k$? In this case the answer is affirmative. 

What about the converse statement? Given a sequence $\{f(\vec{y}_j)\}_{j=1}^{\infty} \in l^p$, does it follow that $f\in L^p(\mathbb{R}^k)$? What if for any sequence of points $\{\vec{y}_j\}_{j=1}^{\infty}$ in the domain of $f$, the sequence  $\{f(\vec{y}_j)\}_{j=1}^{\infty}$ belongs to $l^p$? What if the measure $\mu$ of the domain $\mathcal{D}$ of $\mathfrak{f}$ is not a Lebesgue measure?

 Things get even more bizarre if the set function $\mu$ is only finitely additive. In some cases  $L^p$ spaces may not be complete for any $p>0$. In Measure Theory, "functions" that agree almost everywhere are indistinguishable. In case of finitely additive measures the equivalence classes of a "function" are often  more complex.  Why would anyone care about finitely (and not countably) additive measure  on $\mathcal{D}$? From the point of view of Constructive Mathematics, it is impossible to verify countable additivity of $\mu$ in the first place.

\vspace{5mm}

\section{Some Considerations of feed forward neural networks}

Using a gradient descent method, also known as back-propagation, to optimize weights $\vec{w_j}$, one obtains critical points of the function  $\mathfrak{C} = \norm{\mathfrak{f}-\tilde{f}}_{d}$. Statistically speaking, saddle points in $\mathbb{R}^n$ are more probable than maxima and minima if $n>2$. Thus clever enough gradient descent algorithm will not yield a saddle point. There is no guaranty that the function $\mathfrak{C}$ does not have more than one local minimum, in which case such algorithm may converge to a local  instead of the global minimum of $\mathfrak{C}$. 

%A gradient descent could be used if $\tilde{f}$ is differentiable almost everywhere, so one may choose various characteristic and "relu" type functions as a layer activation function $P_i$.

It is also important to know weather a gradient descent algorithm converges to a local minimum of $\mathfrak{C}$. One may think that should not be a problem. Unfortunately most variable learning rate algorithms (keras optimizers) are designed to increase the convergence rate without a guarantee of asymptotic convergence to a local minimum. The are a few computational problems with  such algorithms. One problem comes from the fact that all weights are updated simultaneously after each instance (epoch) which may cause the subsequent set of weights to yield a larger value of $\mathfrak{C}$. Another problem arises from the choice of the learning rate $\norm{\triangle \vec{w}_j}_2$ for each instance. It is often  proportional to the absolute value of the gradient of $P_j$.  If function $\mathfrak{C}$ is concave up near   local minimum, the value $\norm{\triangle \vec{w}_j}_2$ is likely to be too large.

Another common practice in supervised machine learning is to partition given labeled data set into the training and validation subsets. The validation set is only used to estimate the accuracy of the model at each stage. The goal of the algorithm is to minimize both training and validation error. This idea implicitly assumes some similarity of the function $\mathfrak{f}$  on these two sets.

\begin{wrapfigure}{r}{0.5\textwidth}
    \centering
    \includegraphics[width=0.5\textwidth]{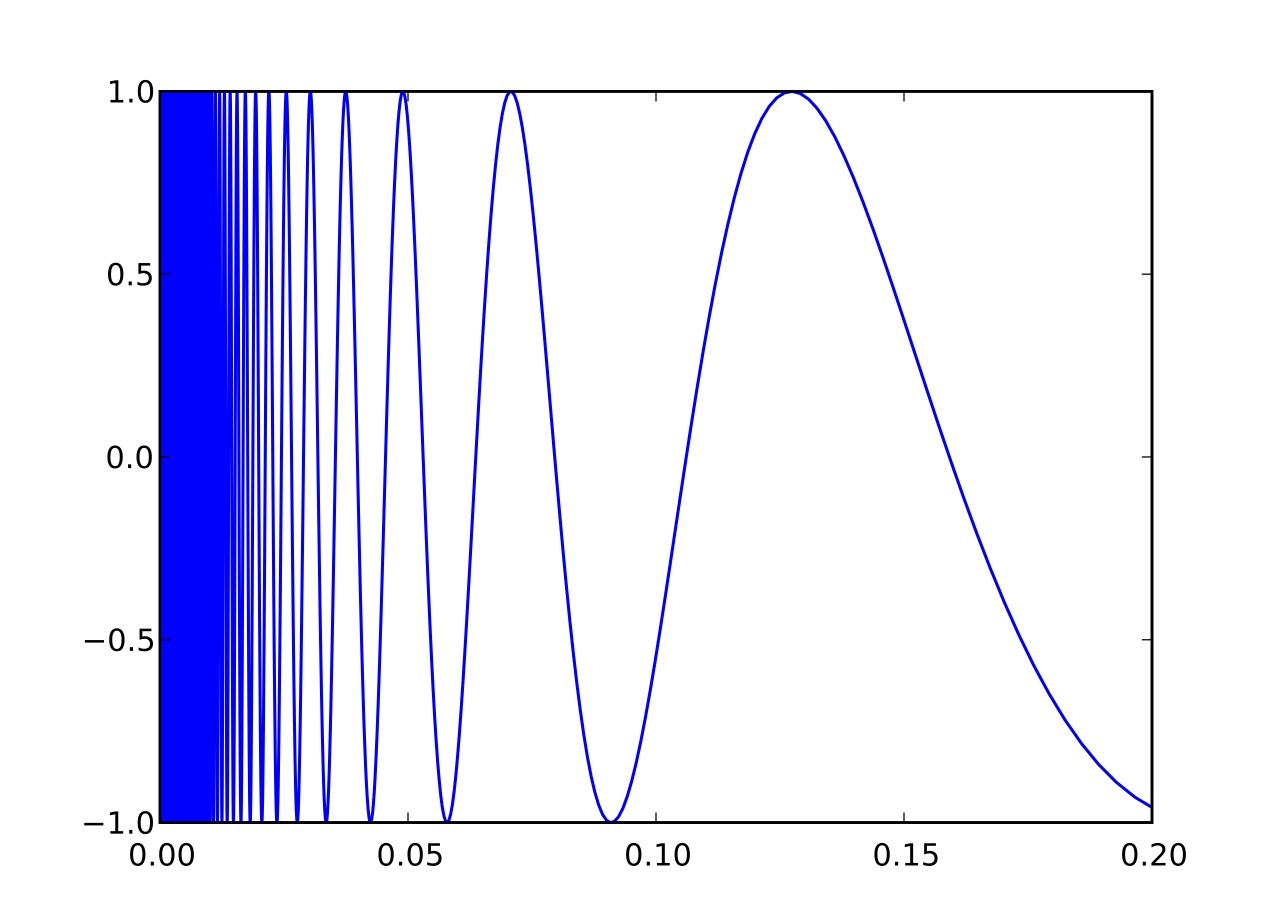}
\end{wrapfigure}
Assume, for example, we are trying to approximate the function
\begin{equation}
\mathfrak{f}(x) = \sin\left(\frac{1}{x} \right), \qquad x>0
\end{equation}

Try to approximate this function using a neural network of any size so that the mean square error on $10^3$ random points of the interval $(0, \, 0.01)$ is less than 0.1. Using Tensorflow 2 with four fully connected layers of size 800, activation function LeakyReLU(alpha=0.01), on $10^5$  randomly generated points in the interval $(0.001, \, 0.011)$ after $10^4$ epochs with Adam optimizer and validation split $0.2$ I obtained the following  approximation and mean square error.

\begin{figure}[H]
	\centering
	\subfloat[]{%
		\label{subfig1}%
		\includegraphics[width=.4\linewidth]{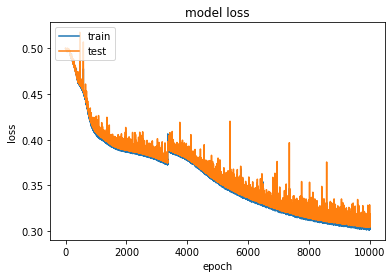}%
	}%
	\qquad
	\subfloat[]{%
		\label{subfig2}%
		\includegraphics[width=.4\linewidth]{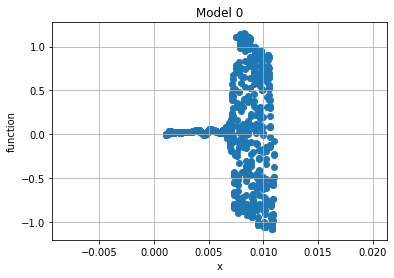}%
	}
	
	\caption{Model History and Evaluation \protect\subref{subfig1} and \protect\subref{subfig2}.}
	\label{myfig0}
	
\end{figure}

With with sixteen fully connected layers of size 200 and other parameters unchanged, the  approximation and mean square error are as follows.

\begin{figure}[H]
	\centering
	\subfloat[]{%
		\label{subfig1}%
		\includegraphics[width=.4\linewidth]{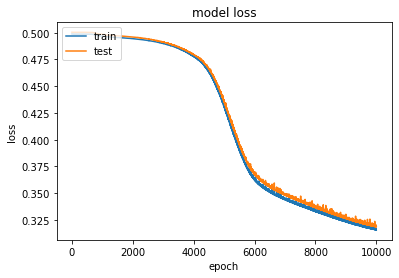}%
	}%
	\qquad
	\subfloat[]{%
		\label{subfig2}%
		\includegraphics[width=.4\linewidth]{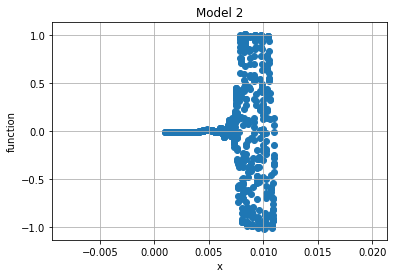}%
	}
	
	\caption{Model History and Evaluation \protect\subref{subfig1} and \protect\subref{subfig2}.}
	\label{myfig2}
	
\end{figure}

The following example is somewhat challenging, but even more pathological.
\begin{example}
\label{example:1}
	One can construct a measurable subset $S$ of $[0,1]$ with the following properties. Both $S$ and its complement, $S^c$, in $[0, 1]$ are totally disconnected (contain no interval) and
	\begin{equation}
	\mu(S)=\mu(S^c)=\frac{1}{2}.
	\end{equation}
	Let $\mathfrak{f}$ be the characteristic function of such set,
	\begin{equation}
	\mathfrak{f}(x)=I_{S}(x), \qquad x\in [0, 1].
	\end{equation} 
	Approximation of $\mathfrak{f}$ by a  function $\tilde{f}$ in (\ref{eq:modnn3}) on any infinite countable subset of $[0,1]$ with respect to $l^2$ norm is computationally impossible because for each $x\in [0, 1]$ the probability of $\{f(x)=0\}$ is equal the probability of $\{f(x)=1\}$.
\end{example}

Each domain $\mathcal{D}$ of the above examples is one dimensional, and it is well known that approximation problem becomes more challenging as the dimension increases.

\vspace{7mm}

\section{Simple functions and Adaptive Neural Networks}

Let us take another look at the machine learning problem of approximation of a function $\mathfrak{f}$ given its values at a countable set of points $S=\{\vec{x}_1, \vec{x}_2, ...\}\subset \mathcal{D}\subset \mathbb{R}^k$. If we only assume that $\mathfrak{f}\in L^p(\mathcal{D})$, it may be impossible to  approximate $\mathfrak{f}$ on $S$ with respect to mean square error (see example \ref{example:1}). On the other hand, if we assume some smoothness conditions (like bounded variation), to predict the value of $\mathfrak{f}$ at $\vec{x}\in \mathcal{D}\setminus S$, one could  take the average of values  of $\mathfrak{f}$ in $S$ in some neighborhood of $\vec{x}$. In such case, do we really need deep neural networks to construct $\tilde{f}$ approximation of $\mathfrak{f}$, or is it just an excuse to own the latest Nvidia graphics card?

Note that the current methods in neural networks are using approximation of a given measurable function $\mathfrak{f}$ by almost everywhere continuous function $\tilde{f}$. In Measure Theory one first approximates a measurable function $\mathfrak{f}$ by  simple functions. The whole point of Lebesgue integration is to use  simple functions instead of piece-wise continuous (step) functions. Neural networks representing simple functions would not consume as much computational power required by matrix multiplication.
Some work in this direction \cite{isermann} is known as Lookup Tables, but no   connection between simple functions and lookup tables have been made explicitly.

Another interesting direction would be an algorithm which  designs the structure of a neural network. To implement this approach, such algorithm would have to control the width of layers and the depth of the  network.

To address  the width control, one could partition each layer, $i$, of neurons  into two subsets, $A_i$ and $B_i$, with zero weights of neurons in $A_i$, and non-zero  weights of neurons in $B_i$. Here is where the condition 
\begin{equation}
\label{eq:modnn8}
P_i(\vec{0})=0
\end{equation}
of an activation function $P_i$ becomes significant.  If (\ref{eq:modnn8}) holds, the neurons from $A_i$ contribute nothing to the value of $\tilde{f}$. One could think of neurons in $A_i$ as auxiliary neurons. As long as  weight optimization algorithm does not compute gradients of neurons in $A_i$, the computational complexity of the network is equivalent to one containing  only the neurons in $B_i$'s.

To adapt the depth of a neural network, one could partition all layers of neurons into two sets, $S$ and $T$, such that the layers in $S$ precede the layers in $T$. If  all but one neurons in layer $i$  from $T$ belongs to $A_i$, and  the remaining neuron in $B_i$ assigned $\vec{w}_i$ consisting of ones, as long as  $P_i(x)=x$,  such layer acts as an identity function. If one does not compute the gradients and  update the values $\vec{w}_i$ for layer $i$ in $T$, the computational burden of such layers is negligible.

In other words, the layers in set $T$ are dormant and act as the identity function and each neuron  that belongs to set  $A_i$ acts as a place holder. Increasing  the size of $A_i$'s and  size  $T$  would increases the maximum approximation accuracy with little  increase in computations.

One could then implement a rule for a "switch" of  a layer of neurons in $T$ to a layer in $S$ based on a threshold  of the gradient of  $\mathfrak{C}$. One could similarly implement a rule for a "switch" of  neurons from $A_i$ to $B_i$. Implementation of such switches   automates the growth of width and depth of neural networks to accommodate the approximation accuracy  without any human interference. 

The above described adaptive algorithm is somewhat similar to a learning process of an adult human brain. Such brain  contains constant number of neurons, but the number of neural connections is changing while learning a new skill. One could also consider a reverse switch from class $B_i$ to $A_i$ to implement the ability to "forget" no longer needed skill. Going a step further, one could allow the neurons in each $A_i$ to be shared by multiple networks working in parallel.

\vspace{5mm}

\section{Computer Vision}

\quad One of the promising areas of neural network application is so called computer vision. In recent years convolutional neural networks had a significant progress in object detection and recognition from images and video data. One of the challenges of object classification is a consequence of so-called "curse of dimensionality". Each $m\times n$ pixels image of an object is often treated as $(m\cdot n)$-dimensional vector. Even a moderate size picture of  $1024\times 1024$  pixels without some reprocessing  presents a computational challenge for modern machines 

There is another problem with the idea of representing $m\times n$ pixels images as $(m\cdot n)$-dimensional vectors. By converting 2D objects into  vectors, one loses the internal structure of the underlying Cartesian space. Assume, for example, we have a gray scale $n\times n$ pixel image of a single object $O$. Let $f(x, y)\in [0, 1]$ be the grayness intensity value the pixel with Cartesian coordinates $(x, y)$, with $x, y =1, 2, ... , n$. If $\epsilon>0$ is small, one can usually assume that  functions 
\begin{equation}
\label{eq:modnn50}
f(x\pm 1, y), \quad f(x, y\pm 1), \quad f(n-x, y), \quad f(x, n-y), \quad f(x-1, y), \quad f(x, y)\pm \epsilon 
\end{equation}
which correspond to shifts, reflections and intensity change, would also represent the same object $O$. One could similarly define a small rotation  and noise invariance of the function representation of the given object. Enforcing such invariance rules on the structure of neural networks is far more difficult.

 We think of objects as three dimensional, so one could assume that  dimension of the solution space of object classification should be of same order of magnitude. Perhaps some difficulties of object classification follow from the complexity of equivalence relation defining each class of objects. Often times such relation is not based only on the three dimensional space. How, for example,  would you recognize a bottle? Since bottles come in various shapes sizes, even if we had a rigorous definition of shape, equivalence rule of the "bottle" class would be complicated. Yet, when an adult human is presented with a previously unseen and even unusual bottle, would recognise it without much effort. Our notion of bottles comes from their extensive use  in daily life. It seems unlikely that the problem of computer vision would have a computationally feasible solution by a narrow AI algorithm trained only  on images.

\vspace{5mm}

\section{Recurrent Neural Networks}

 Another class of networks, called recurrent  NNs, are designed to predict the n-the value, $\vec{y}_n$, of a sequence of vectors $\{\vec{y}_n\}_{n\in\mathbb{N}}$, given all previous values $\vec{y}_1, \dots , \vec{y}_{n-1}$. One would think this type of problem  requires different approach, yet the common practice is to modify the connections of feed forward neural network and use good old gradient descent.

 Fourier series was  fist thing came  to my mind when looking at the above problem. If the sequence is periodic we would discover this fact within two periods of the sequence. Since not all functions are periodic, one could next assume the sequence function is almost periodic. For example, the class of Besicovich almost periodic functions on $\mathbb{C}$ consist of trigonometric polynomials of the form
\begin{equation}
\label{eq:pap1}
P(x)=\sum_{k=1}^{n} a(\eta_k) e^{i\eta_k x}, \qquad \eta_k\in \mathbb{R}, \quad a(\eta_k; P)\in \mathbb{C}
\end{equation}
and their completion  with the norm
\begin{equation}
\label{eq:pap2}
\norm{P}_{B^2_{ap}}=\lim_{\tau\to \infty}\left(\frac{1}{2\tau}\int_{-\tau}^{\tau} |P(x)|^2 dx\right)^{1/2}. 
\end{equation}
This is a large set of functions that need not be periodic. Even very easy almost periodic function $f(x)= e^{ix}+ e^{i\sqrt{2} x}$ is not periodic. It would be interesting to use recurrent neural networks to approximate this function.

One can easily generalize Fourier series to this class of functions and use a computational power to estimate the Fourier series  instead of using gradient descent. Since
\begin{equation}
\langle e^{i\eta_jx}, e^{i\eta_kx} \rangle=\lim_{\tau\to \infty}\left(\frac{1}{2\tau}\int_{-\tau}^{\tau} e^{i\eta_jx}\cdot e^{-i\eta_kx}dx\right)=\begin{cases} 
1 & \eta_j=\eta_k \\
0 & \eta_j\ne \eta_k
\end{cases}
\end{equation}
the set of functions functions $\{ e^{i\eta_kx}\}_{k\in\mathbb{N}}$ form an orthonormal system and the completion of trigonometric polynomials in (\ref{eq:pap1}) is a Hilbert space. One may compute generalized  Fourier coefficients of a Besicovich almost periodic function $f$ using
\begin{equation}
\label{eq:pap2a}
a(\eta; f)=\lim_{\tau\to \infty}\left(\frac{1}{2\tau}\int_{-\tau}^{\tau} f(x)e^{-i\eta x} dx\right), \qquad \eta \in \mathbb{R}.
\end{equation}
Moreover, one can take advantage of Harmonic Analysis theory, by first defining a finitely additive measure $\gamma$ with
\begin{equation}
\gamma(S)=\lim_{\tau\to \infty}\left(\frac{1}{2\tau}\int_{(-\tau,  \tau)\cap S} 1 \, dx \right), \qquad S\subset \mathbb{R}
\end{equation}
and then obtaining $\norm{\cdot}_{B^2_{ap}}$ in (\ref{eq:pap2}) as the usual $L^2$ norm with the measure $\gamma$.

Only recently Fourier series were used in the new design of recurrent neural network called  "Transformers" introduced by authors of the paper "Attention is all you need"  \cite{vaswani}.

\end{document}